\documentclass[11pt,twocolumn]{article}

\usepackage[utf8]{inputenc}
\usepackage[T1]{fontenc}
\usepackage{lmodern}
\usepackage{microtype}
\usepackage{amsmath,amssymb,amsfonts}
\usepackage{bm}
\usepackage{graphicx}
\usepackage{booktabs}
\usepackage{tabularx}
\usepackage{float}
\usepackage{caption}
\usepackage[margin=1in]{geometry}
\usepackage{enumitem}
\usepackage{url}
\usepackage[colorlinks=true,linkcolor=blue,citecolor=blue,urlcolor=blue]{hyperref}

\newenvironment{linenomath}{}{}

\newcolumntype{C}{>{\centering\arraybackslash}X}

\title{Evaluating Factor-Wise Auxiliary Dynamics Supervision\\for Latent Structure and Robustness\\in Simulated Humanoid Locomotion}

\author{
Chayanin Chamachot\thanks{Correspondence: 6633041021@student.chula.ac.th \quad ORCID: \href{https://orcid.org/0009-0009-1605-7988}{0009-0009-1605-7988}}\\
Department of Computer Engineering, Faculty of Engineering,\\
Chulalongkorn University, Bangkok 10330, Thailand
}

\date{}

\begin{document}
\maketitle

\begin{abstract}
We evaluate whether factor-wise auxiliary dynamics supervision produces useful latent structure or improved robustness in simulated humanoid locomotion. DynaMITE---a transformer encoder with a factored 24-d latent trained by per-factor auxiliary losses during proximal policy optimization (PPO)---is compared against Long Short-Term Memory (LSTM), plain Transformer, and Multilayer Perceptron (MLP) baselines on a Unitree G1 humanoid across four Isaac Lab tasks. The supervised latent shows no evidence of decodable or functionally separable factor structure: probe $R^2 \approx 0$ for all five dynamics factors, clamping any subspace changes reward by $< 0.05$, and standard disentanglement metrics (MIG, DCI, SAP) are near zero. An unsupervised LSTM hidden state achieves higher probe $R^2$ (up to 0.10). A $2 \times 2$ factorial ablation ($n = 10$ seeds) isolates the contributions of the $\tanh$ bottleneck and auxiliary losses: the auxiliary losses show no measurable effect on either in-distribution (ID) reward ($+0.03$, $p = 0.732$) or severe out-of-distribution (OOD) reward ($+0.03$, $p = 0.669$), while the bottleneck shows a small, consistent advantage in both regimes (ID: $+0.16$, $p = 0.207$; OOD: $+0.10$, $p = 0.208$). The bottleneck advantage persists under severe combined perturbation but does not amplify, indicating a training-time representation benefit rather than a robustness mechanism. LSTM achieves the best nominal reward on all four tasks ($p < 0.03$); DynaMITE degrades less under combined-shift stress (2.3\% vs.\ 16.7\%), but this difference is attributable to the bottleneck compression, not the auxiliary supervision. For locomotion practitioners: auxiliary dynamics supervision does not produce an interpretable estimator and does not measurably improve reward or robustness beyond what the bottleneck alone provides; recurrent baselines remain the stronger choice for nominal performance.
\end{abstract}

\noindent\textbf{Keywords:} reinforcement learning; humanoid locomotion; domain randomization; latent dynamics; auxiliary losses; sim-to-real; disentanglement; representation analysis

\vspace{0.5em}

\section{Introduction}\label{sec:intro}

Reinforcement learning (RL) policies for humanoid locomotion are typically trained in simulation with domain randomization (DR)~\cite{dr_tobin,dr_openai,dr_anymal} and must transfer to conditions that differ from the training distribution. One strategy for handling dynamics variation is latent dynamics identification: training an encoder to infer a compact latent representation of the environment's physical parameters from an observation--action history, then conditioning the policy on this estimate. Rapid Motor Adaptation (RMA)~\cite{rma} demonstrated this approach with a two-phase training pipeline, achieving sim-to-real transfer on a quadruped. Subsequent work~\cite{rma_followup,drac} has extended the paradigm, generally reporting improved robustness.

A natural refinement is to supervise the latent representation per factor---training dedicated sub-vectors to predict individual dynamics parameters (friction, mass, motor strength, etc.) rather than a monolithic estimate. The intuition is appealing: a factored, decodable latent should enable both better generalization and post-hoc interpretability, allowing practitioners to inspect \emph{what} the policy believes about each physical quantity.

We test this intuition rigorously. Our architecture, DynaMITE (Dynamics-Matching Inference via Transformer Encoding), maps a short history through a transformer encoder to a 24-dimensional factored latent with five dedicated subspaces trained by per-factor auxiliary losses during PPO. We compare it against LSTM, plain Transformer, and MLP baselines on a Unitree G1 humanoid across four locomotion tasks in NVIDIA Isaac Lab, using 5 seeds for the main comparison, 10 seeds for ablations, and over 7,000 evaluation episodes for push recovery.

The central finding is negative: factor-wise auxiliary dynamics supervision does not produce a decodable or functionally separable latent representation in this setting. Probe $R^2 \approx 0$ for all five factors (linear and MLP probes); clamping any factor subspace changes reward by less than 0.05; and standard disentanglement metrics (MIG, DCI, SAP) are near zero. An unsupervised LSTM hidden state achieves higher probe $R^2$ (up to 0.10). A $2 \times 2$ factorial ablation isolates the bottleneck and auxiliary-loss contributions: the auxiliary losses show no measurable effect ($p > 0.66$), while the $\tanh$ bottleneck shows a small, consistent advantage ($p \approx 0.2$). DynaMITE degrades less than LSTM under severe combined-shift perturbation (2.3\% vs.\ 16.7\%), but the factorial confirms this reflects the bottleneck compression, not the auxiliary supervision.

Our analysis uses six complementary tools: probes (decodability), interventions (functional separability), gradient flow (training dynamics), SVD geometry (representation structure), KNN-based MI (information content), and standard disentanglement metrics (MIG, DCI, SAP). Across all six, results for dynamics identification are null or weak. We report these negative findings in full, following calls for more complete reporting of null results in RL~\cite{publication_bias}.

\section{Related Work}\label{sec:related}

\subsection{Domain Randomization and Sim-to-Real Transfer}

Domain randomization (DR) trains policies over distributions of simulator parameters to improve robustness at deployment~\cite{dr_tobin,dr_openai}. Successes span quadrupeds~\cite{dr_anymal,dr_a1,dr_solo}, bipeds~\cite{dr_humanoid_1,dr_humanoid_2,dr_cassie}, and full humanoids~\cite{dr_humanoid_full_1,dr_humanoid_full_2}. However, excessively wide randomization can degrade nominal performance~\cite{dr_limitations,dr_conservative,dr_width_tradeoff}, motivating adaptive methods that infer dynamics online rather than relying solely on broad randomization.

\subsection{History-Conditioned Policies}

LSTM-based policies implicitly adapt via hidden state dynamics~\cite{lstm_locomotion_1,lstm_locomotion_2,lstm_locomotion_3,lstm_adaptation}. Transformers process a fixed-length history window with attention~\cite{transformer_context,transformer_locomotion,dr_humanoid_full_1}, enabling parallel training and explicit context length control. Decision Transformer~\cite{decision_transformer} and Algorithm Distillation~\cite{incontext_rl} frame RL as sequence modeling but target offline and meta-RL respectively; our setting is online PPO with DR.

\subsection{Latent Dynamics Identification}

Classical system identification recovers physics parameters from input--output data~\cite{sysid_classical_1,sysid_classical_2}. RMA~\cite{rma} introduced a two-phase approach: train a privileged teacher with access to ground-truth dynamics, then distill an adaptation module that infers a latent dynamics estimate from history. World models~\cite{latent_dynamics_vae_1,latent_dynamics_vae_2} learn latent dynamics for planning; context-conditioned policies~\cite{context_conditional_1,context_conditional_2} achieve similar goals in a meta-RL framework. DynaMITE differs by integrating per-factor auxiliary supervision directly into the PPO training loop via a factored bottleneck, avoiding two-phase distillation.

\subsection{Representation Analysis in RL}

Probing classifiers assess what information is decodable from learned representations~\cite{probing_rl_1,probing_rl_2}. Disentanglement metrics (MIG~\cite{mig}, DCI~\cite{dci}, SAP~\cite{sap}) quantify factor alignment in generative models but have seen limited application to RL latent spaces. Recent work in RL representation analysis~\cite{disentanglement_rl} studies zero-shot transfer via disentangled features. Our analysis applies six complementary tools to evaluate whether auxiliary supervision produces the intended latent structure.

\section{Method}\label{sec:method}

\begin{figure}[t]
\centering
\includegraphics[width=\columnwidth]{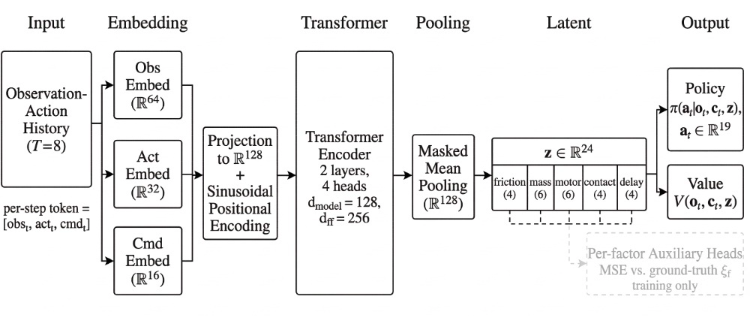}
\caption{Overview of the DynaMITE architecture. A two-layer transformer encoder processes an 8-step (160\,ms) observation--action history to produce a 24-dimensional factored latent vector $\bm{z} \in \mathbb{R}^{24}$, decomposed into five factor subspaces (friction, mass, motor strength, contact stiffness, action delay). Each subspace is trained with a dedicated auxiliary dynamics-prediction loss during PPO training. The latent is concatenated with the current observation and fed to the policy $\pi(a \mid s, \bm{z})$ and value $V(s, \bm{z})$ heads. Auxiliary losses are active only during training.\label{fig:architecture}}
\end{figure}

The DynaMITE architecture (Figure~\ref{fig:architecture}) maps a short observation--action history to a factored latent vector concatenated with the current observation before being fed to the policy and value heads.

\subsection{History Buffer and Transformer Encoder}\label{sec:method_encoder}

At each control step $t$, the agent maintains a history buffer of the $H = 8$ most recent observation--action pairs $\{(o_{t-H+1}, a_{t-H+1}), \ldots, (o_t, a_t)\}$, corresponding to a 160\,ms context window at the 50\,Hz control rate. Each pair is embedded via learned linear projections and summed with sinusoidal positional encodings. The resulting sequence of tokens is processed by a transformer encoder with 2 layers, 4 attention heads, and model dimension $d_\text{model} = 128$. The encoder output is mean-pooled across the sequence dimension to yield a fixed-size context vector $\bm{h} \in \mathbb{R}^{128}$.

\subsection{Factored Latent Head}\label{sec:method_latent}

A linear projection followed by a $\tanh$ nonlinearity maps $\bm{h}$ to a 24-dimensional latent vector $\bm{z} \in \mathbb{R}^{24}$, which is explicitly partitioned into five contiguous factor subspaces: friction (dims 0--3), mass (dims 4--9), motor strength (dims 10--15), contact stiffness (dims 16--19), and action delay (dims 20--23). The subspace sizes are chosen proportional to the expected complexity of each factor. During training, each subspace $\bm{z}_f$ receives a dedicated auxiliary loss that trains it to predict the corresponding ground-truth dynamics parameter $\theta_f$:

\begin{linenomath}
\begin{equation}
\mathcal{L}_{\text{aux},f} = \| g_f(\bm{z}_f) - \theta_f \|^2,
\end{equation}
\end{linenomath}

\noindent where $g_f$ is a small MLP head. The total training loss is:

\begin{linenomath}
\begin{equation}
\mathcal{L} = \mathcal{L}_{\text{PPO}} + c_v \mathcal{L}_{\text{value}} + 0.1 \sum_{f} \mathcal{L}_{\text{aux},f},
\end{equation}
\end{linenomath}

\noindent with auxiliary loss weight fixed at 0.1. Auxiliary losses are active only during training; at evaluation time, the auxiliary heads are discarded.

\subsection{Policy and Value Heads}\label{sec:method_policy}

The latent $\bm{z}$ is concatenated with the current observation $o_t$ and fed to shared-architecture policy and value MLPs. The policy head outputs a Gaussian distribution $\pi(a \mid o_t, \bm{z})$ from which actions are sampled during training; at evaluation, actions are taken as the distribution mean (deterministic mode). The value head outputs $V(o_t, \bm{z})$ for advantage estimation.

\subsection{Baseline Architectures}\label{sec:method_baselines}

All four model architectures share the same observation embedding, action embedding, policy MLP, and value MLP. They differ only in the history encoding mechanism:

\begin{table}[H]
\caption{Model architecture summary. Parameter counts vary by task due to different observation dimensions.\label{tab:architectures}}
\begin{tabularx}{\columnwidth}{CCCCC}
\toprule
\textbf{Model} & \textbf{History} & \textbf{Latent} & \textbf{Aux Loss} & \textbf{Params} \\
\midrule
MLP & None & No & No & 266--362k \\
LSTM & Hidden state & No & No & 176--215k \\
Transformer & 8 steps & No & No & 330--342k \\
DynaMITE & 8 steps & 24-d factored & Yes & 342--392k \\
\bottomrule
\end{tabularx}
\end{table}

The \textbf{MLP} baseline receives only the current observation with no history. The \textbf{LSTM} baseline processes observations sequentially and conditions the policy on its hidden state. The \textbf{Transformer} baseline uses the same encoder as DynaMITE but without the factored latent head or auxiliary losses---the mean-pooled context vector is passed directly to the policy and value heads. This baseline controls for the transformer encoder architecture, but does not isolate the auxiliary losses from the 24-d $\tanh$ bottleneck: DynaMITE adds both the bottleneck and the auxiliary supervision simultaneously, so any behavioral difference between Transformer and DynaMITE may reflect either component or their interaction. The $2 \times 2$ factorial ablation (Section~\ref{sec:results_factorial}) addresses this confound by independently varying bottleneck usage and auxiliary losses; all factorial conditions share identical architecture, training protocol, and evaluation.

\section{Evaluation Protocol}\label{sec:eval_protocol}

To reduce analytic flexibility, we fixed the evaluation protocol---including metrics, sweep levels, and checkpoint-selection criteria---before the main experiment campaign.

\subsection{Training Protocol}\label{sec:eval_training}

All models are trained with PPO (clipped objective, generalized advantage estimation (GAE))~\cite{dr_openai} using 512 parallel Isaac Lab environments. Each training run spans 10 million timesteps at a control frequency of 50\,Hz ($\Delta t = 20$\,ms), requiring approximately 14 minutes on a single NVIDIA RTX 4060 Laptop GPU. Checkpoints are saved every 614,400 steps (${\sim}60$\,s), and the best checkpoint is selected by training-time stochastic evaluation reward. All reported results then use deterministic evaluation (Section~\ref{sec:eval_deterministic}); the stochastic-to-deterministic gap is a potential source of noise but is consistent across all models.

\subsection{Deterministic Evaluation}\label{sec:eval_deterministic}

All reported numbers use \textbf{deterministic evaluation}: actions are taken as the distribution mean with no sampling. Each evaluation consists of 100 episodes (main comparison, ablations) or 50 episodes (OOD sweeps, push recovery). Episodes use fixed-length rollouts with no early termination. The evaluation environment seed is fixed at 42 for all models within a task, ensuring identical initial conditions. Environment resets fully randomize initial joint positions and domain parameters.

The main comparison spans 5 training seeds (42--46) $\times$ 4 tasks $\times$ 4 models $= 80$ evaluations. OOD sweeps cover 5 training seeds $\times$ 4 models $\times$ 5 sweep types $\times$ 3 tasks $= 140$ evaluations. Push recovery evaluates 5 seeds $\times$ 4 models $\times$ 7 magnitudes $\times$ 50 episodes $= 7{,}000$ episodes. Mechanistic analyses use 3 seeds for gradient flow (10M steps each) and 5 seeds for geometry and mutual information estimation (${\sim}36{,}000$ representations per model/seed).

\subsection{Metrics}\label{sec:eval_metrics}

\textbf{Reward.} Penalty-based (always negative); higher (less negative) values indicate better performance. A method achieving $-4.18$ vs.\ $-4.48$ accumulates approximately 6\% less penalty per step.

\textbf{OOD sensitivity} (custom metric). Defined as $\max(\bar{r}) - \min(\bar{r})$ across sweep levels, where $\bar{r}$ is the mean reward at each level. Lower values indicate a flatter reward curve across perturbation levels. \emph{Caveat:} this metric does not distinguish models with uniformly poor performance from models that genuinely resist degradation; a model that performs badly everywhere will also have low sensitivity. It should therefore be interpreted alongside absolute reward levels (severe-level mean, worst-case reward). We also report the severe-level mean (average of the two highest-severity levels), worst-case reward (most negative across all sweep levels), and degradation (absolute difference between ID reward and severe-level mean, expressed also as a percentage of $|\text{ID Reward}|$).

\textbf{Recovery time} (custom metric). Steps from push onset until the velocity tracking error drops below a threshold of 1.5. Measured in the push-recovery protocol with controlled push magnitudes.

\textbf{Factor alignment} (custom metric). A within-factor Pearson correlation ratio, not a standard disentanglement benchmark such as MIG, DCI, or SAP. Chance level is 0.20 for five factors. Not comparable to metrics from prior disentanglement work.

\subsection{Statistical Reporting}\label{sec:eval_stats}

For the main comparison ($n = 5$ seeds), we report mean $\pm$ standard deviation, 95\% confidence intervals ($t$-distribution), and paired $t$-tests. For ablations ($n = 10$ seeds), paired $t$-tests are computed against the full DynaMITE model. For OOD sweeps, Holm--Bonferroni corrected $p$-values are used. Only 3 of 42 pairwise OOD comparisons survive correction at $p_\text{adj} < 0.05$; most ranking claims are therefore directional rather than statistically confirmed.

\section{Results}\label{sec:results}

\subsection{Nominal In-Distribution Comparison}\label{sec:results_id}

\begin{figure}[t]
\centering
\includegraphics[width=\columnwidth]{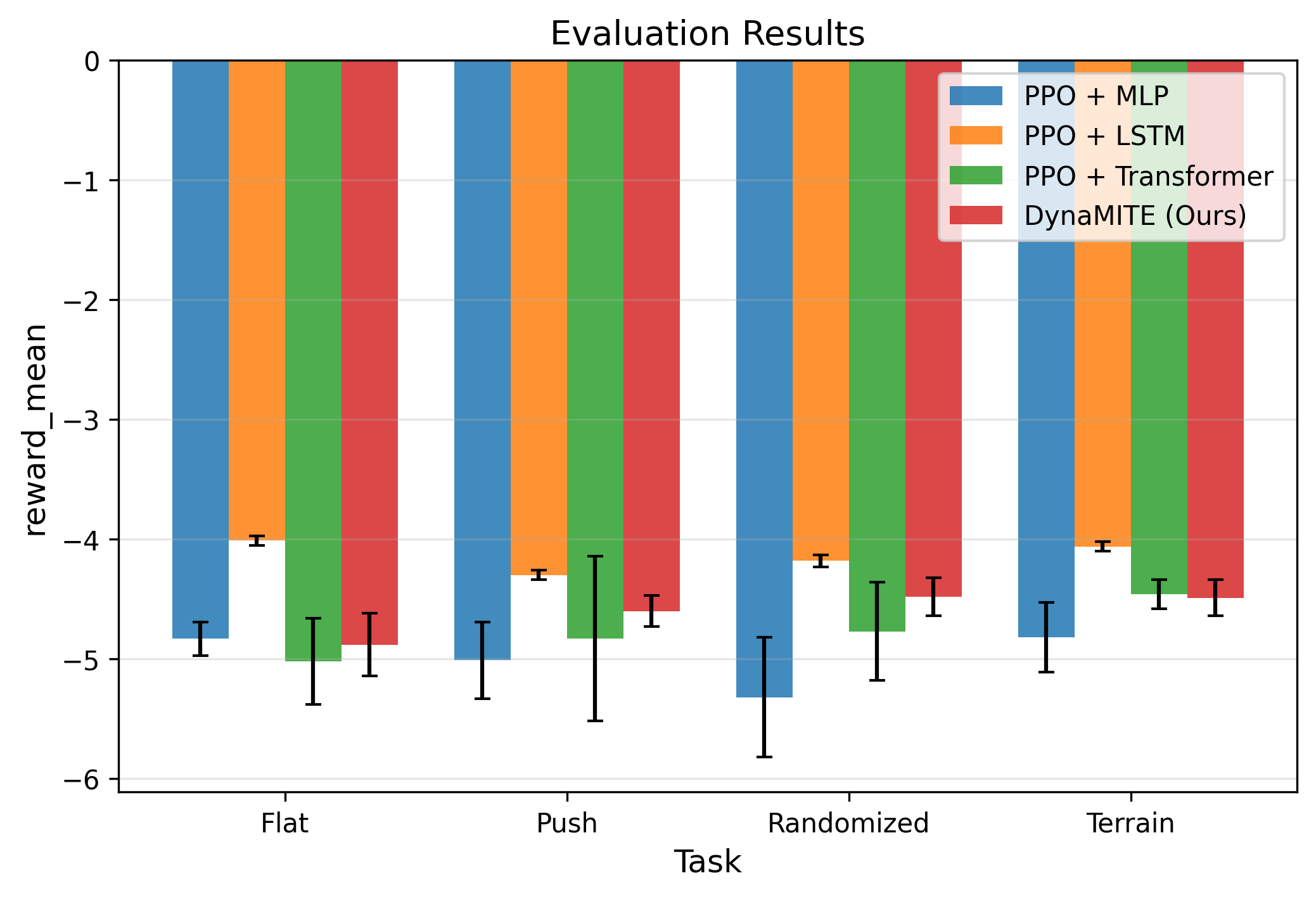}
\caption{In-distribution reward (5 seeds, deterministic evaluation). LSTM achieves the best reward on all tasks; DynaMITE ranks second on three of four tasks but is significantly worse than LSTM on all.\label{fig:eval_bars}}
\end{figure}

Table~\ref{tab:id_comparison} presents the mean $\pm$ standard deviation of cumulative reward across five training seeds and four tasks under deterministic 100-episode evaluation. LSTM achieves the best reward on all four tasks with the lowest variance ($p < 0.03$, paired $t$-test; Table~\ref{tab:paired_tests}). DynaMITE ranks second on three of four tasks (Figure~\ref{fig:eval_bars}).

\begin{table}[H]
\caption{In-distribution reward comparison (5 seeds, deterministic eval). Bold indicates best per task. All LSTM vs.\ DynaMITE paired $t$-tests are significant ($p < 0.03$).\label{tab:id_comparison}}
\small
\begin{tabularx}{\columnwidth}{CCCCC}
\toprule
\textbf{Method} & \textbf{Flat} & \textbf{Push} & \textbf{Rand.} & \textbf{Terrain} \\
\midrule
MLP & $-4.83$ & $-5.01$ & $-5.32$ & $-4.82$ \\
LSTM & $\mathbf{-4.01}$ & $\mathbf{-4.30}$ & $\mathbf{-4.18}$ & $\mathbf{-4.06}$ \\
Transf. & $-5.02$ & $-4.83$ & $-4.77$ & $-4.46$ \\
DynaMITE & $-4.88$ & $-4.60$ & $-4.48$ & $-4.49$ \\
\bottomrule
\end{tabularx}
\end{table}

\begin{table}[H]
\caption{Paired $t$-tests: LSTM vs.\ DynaMITE (matched training seeds, $n = 5$).\label{tab:paired_tests}}
\begin{tabularx}{\columnwidth}{CCCC}
\toprule
\textbf{Task} & \textbf{Mean Diff} & \textbf{$t$} & \textbf{$p$} \\
\midrule
Flat & $+0.870$ & 8.49 & 0.0011 \\
Push & $+0.305$ & 4.69 & 0.0094 \\
Randomized & $+0.303$ & 3.34 & 0.029 \\
Terrain & $+0.435$ & 8.55 & 0.0010 \\
\bottomrule
\end{tabularx}
\end{table}

\subsection{Combined-Shift Stress Test}\label{sec:results_combined}

To simulate the compounding dynamics mismatch characteristic of sim-to-real transfer, we simultaneously shift friction, push magnitude, and action delay across five severity levels. Table~\ref{tab:combined_reward} reports reward; Table~\ref{tab:combined_tracking} reports tracking error.

\begin{table}[H]
\caption{Combined-shift reward across severity levels (randomized task, 5 seeds, 50-episode deterministic eval per level). Bold indicates best at each level.\label{tab:combined_reward}}
\small
\begin{tabularx}{\columnwidth}{CCCCC}
\toprule
\textbf{Method} & \textbf{L0 (ID)} & \textbf{L1} & \textbf{L2} & \textbf{L4} \\
\midrule
DynaMITE & $-4.38$ & $-4.47$ & $-4.50$ & $\mathbf{-4.63}$ \\
LSTM & $\mathbf{-3.56}$ & $\mathbf{-4.23}$ & $\mathbf{-4.40}$ & $-5.12$ \\
Transf. & $-4.67$ & $-4.73$ & $-4.72$ & $-4.86$ \\
MLP & $-5.68$ & $-5.61$ & $-5.55$ & $-5.64$ \\
\bottomrule
\end{tabularx}
\end{table}

\begin{table}[H]
\caption{Combined-shift tracking error across severity levels (randomized task, 5 seeds, 50-episode deterministic eval per level).\label{tab:combined_tracking}}
\small
\resizebox{\columnwidth}{!}{%
\begin{tabular}{lcccccc}
\toprule
\textbf{Method} & \textbf{L0} & \textbf{L1} & \textbf{L2} & \textbf{L3} & \textbf{L4} \\
\midrule
DynaMITE & $2.15$ & $2.56$ & $3.18$ & $4.23$ & $6.13$ \\
LSTM & $\mathbf{1.31}$ & $\mathbf{2.19}$ & $\mathbf{3.01}$ & $4.59$ & $6.94$ \\
Transf. & $2.08$ & $2.58$ & $3.17$ & $4.35$ & $6.48$ \\
MLP & $2.24$ & $2.76$ & $3.40$ & $4.47$ & $6.30$ \\
\bottomrule
\end{tabular}%
}
\end{table}

\begin{figure}[t]
\centering
\includegraphics[width=\columnwidth]{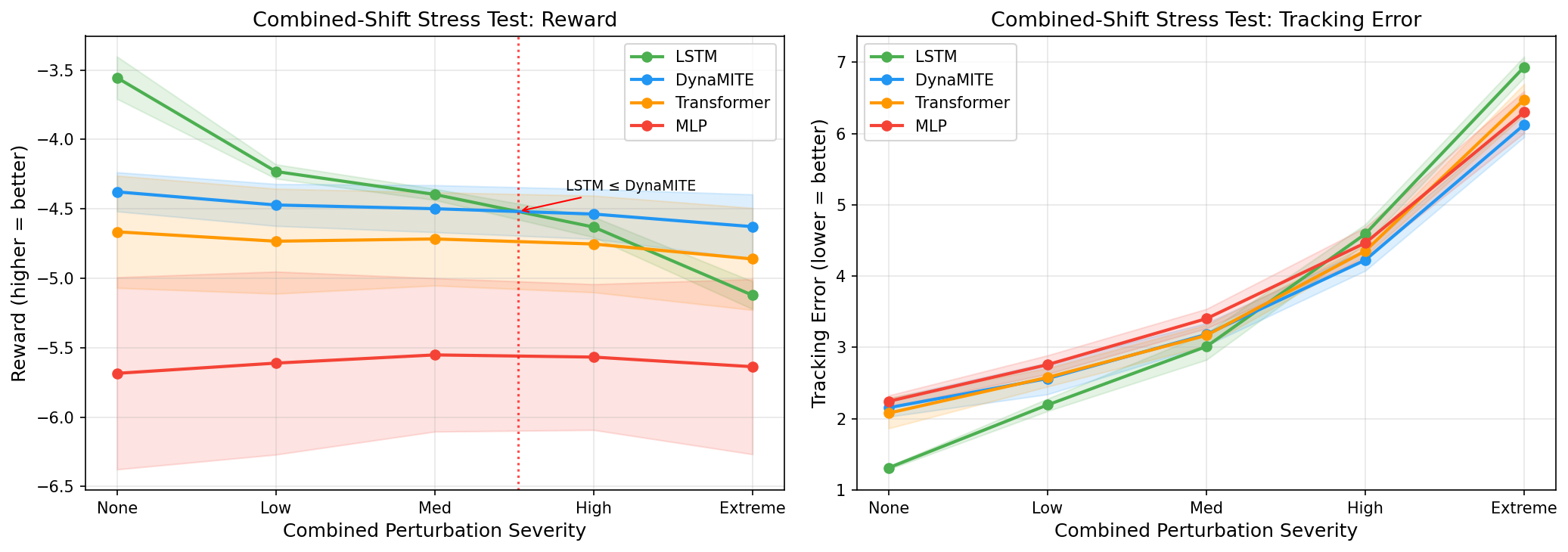}
\caption{Combined-shift stress test (randomized task, 5 seeds). LSTM achieves the best reward at low severity but degrades steeply; DynaMITE's reward is lower at baseline but more stable. The crossover occurs at severity level~3. Neither model dominates across all levels.\label{fig:combined_shift}}
\end{figure}

Figure~\ref{fig:combined_shift} and Table~\ref{tab:severe_ood} quantify severe OOD degradation. Among the two highest-performing ID models, LSTM degrades substantially more than DynaMITE under combined shift (16.7\% vs.\ 2.3\% relative to ID reward). DynaMITE's mean reward matches or exceeds LSTM's from severity level~3 onward; at level~4, LSTM's tracking error (6.94) is the highest among all models, reversing its low-perturbation advantage. Note, however, that the Transformer and MLP baselines also show low sensitivity values (0.20 and 0.13, respectively)---in MLP's case because it performs poorly across all levels rather than because it resists degradation.

Note that the combined-shift Level~0 rewards (e.g., LSTM $-3.56$) differ from the main-comparison ID rewards in Table~\ref{tab:id_comparison} (LSTM $-4.18$): Level~0 pins all dynamics to nominal values (friction $= 1.0$, no push, no delay), while the main comparison evaluates under the full domain randomization distribution. LSTM benefits disproportionately from nominal conditions ($\Delta = 0.62$) compared with DynaMITE ($\Delta = 0.10$), which itself is consistent with higher sensitivity to dynamics variation.

\begin{table}[H]
\caption{Severe OOD degradation (combined-shift sweep, randomized task, 5 seeds, deterministic eval). Degradation $= |\text{Severe} - \text{ID}|$; percentages relative to $|\text{ID}|$.\label{tab:severe_ood}}
\small
\begin{tabularx}{\columnwidth}{CCCCC}
\toprule
\textbf{Model} & \textbf{ID} & \textbf{Severe} & \textbf{Worst} & \textbf{Degrad.} \\
\midrule
DynaMITE & $-4.48$ & $-4.58$ & $-4.63$ & $\mathbf{2.3\%}$ \\
LSTM & $-4.18$ & $-4.88$ & $-5.12$ & $16.7\%$ \\
Transf. & $-4.77$ & $-4.81$ & $-4.86$ & $0.8\%$ \\
MLP & $-5.32$ & $-5.66$ & $-5.68$ & $6.4\%$ \\
\bottomrule
\end{tabularx}
\end{table}

\subsection{Pareto Analysis}\label{sec:results_pareto}

\begin{figure}[t]
\centering
\includegraphics[width=0.85\columnwidth]{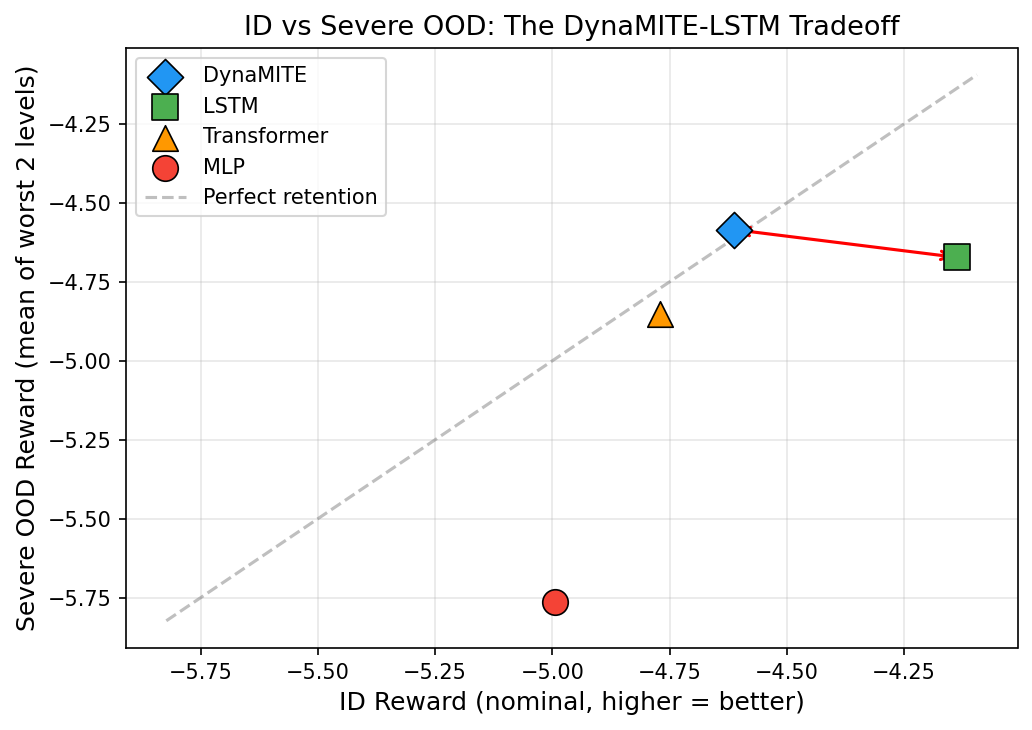}
\caption{Pareto front: in-distribution reward vs.\ severe OOD reward. No model dominates both axes. LSTM achieves the best ID reward; DynaMITE has the highest mean severe OOD reward.\label{fig:pareto}}
\end{figure}

Aggregating across randomized-task OOD sweeps (combined-shift, push magnitude, friction), Table~\ref{tab:pareto} and Figure~\ref{fig:pareto} show that DynaMITE has the highest mean severe OOD reward in our sample ($-4.58$) and best worst-case ($-4.78$), while LSTM achieves the best ID reward ($-4.14$). No model dominates both axes under our protocol.

\begin{table}[H]
\caption{Pareto analysis (5 seeds, deterministic eval): ID reward $=$ 4-task mean; Severe OOD Mean $=$ average across three randomized-task sweeps.\label{tab:pareto}}
\begin{tabularx}{\columnwidth}{CCCC}
\toprule
\textbf{Model} & \textbf{ID} & \textbf{Severe OOD} & \textbf{Worst} \\
\midrule
DynaMITE & $-4.61$ & $\mathbf{-4.58}$ & $\mathbf{-4.78}$ \\
LSTM & $\mathbf{-4.14}$ & $-4.67$ & $-5.12$ \\
Transf. & $-4.77$ & $-4.85$ & $-5.08$ \\
MLP & $-5.00$ & $-5.76$ & $-5.97$ \\
\bottomrule
\end{tabularx}
\end{table}

\subsection{Push Recovery}\label{sec:results_push}

In a controlled push-recovery protocol on flat terrain, the robot walks for 30 steps (steady-state), receives an exact-magnitude push in a random direction, and recovers over 40 steps. Recovery is defined as the tracking error returning below 1.5. Each condition uses 50 episodes per magnitude per seed across 5 seeds. Table~\ref{tab:recovery_steps} reports recovery time; Table~\ref{tab:recovery_peak} reports peak tracking error.

\begin{table}[H]
\caption{Steps to recover command tracking after push (flat task, 5 seeds $\times$ 50 episodes per magnitude). Bold indicates fastest.\label{tab:recovery_steps}}
\small
\begin{tabularx}{\columnwidth}{CCCCC}
\toprule
\textbf{Push} & \textbf{DynaMITE} & \textbf{LSTM} & \textbf{Transf.} & \textbf{MLP} \\
\midrule
1.0 & $\mathbf{5.6}$ & $9.2$ & $6.7$ & $6.3$ \\
2.0 & $\mathbf{5.9}$ & $15.9$ & $8.1$ & $7.6$ \\
3.0 & $\mathbf{6.0}$ & $20.6$ & $8.3$ & $7.8$ \\
4.0 & $\mathbf{6.0}$ & $19.7$ & $8.3$ & $8.1$ \\
5.0 & $\mathbf{6.1}$ & $19.8$ & $8.0$ & $8.2$ \\
6.0 & $\mathbf{6.1}$ & $19.5$ & $8.1$ & $8.7$ \\
8.0 & $\mathbf{6.2}$ & $19.6$ & $8.3$ & $8.3$ \\
\bottomrule
\end{tabularx}
\end{table}

DynaMITE returns below the tracking-error threshold (1.5) in approximately 6 steps regardless of push magnitude (5.6--6.2, nearly constant). LSTM requires 9 to 20 steps to reach the same threshold as push magnitude grows, then plateaus; it is the slowest of all four models under this metric. At pushes $\geq 3$\,m/s, DynaMITE crosses the threshold $3.4\times$ faster than LSTM.

\begin{table}[H]
\caption{Peak tracking error after push (flat task, 5 seeds $\times$ 50 episodes). Bold indicates lower peak error.\label{tab:recovery_peak}}
\begin{tabularx}{\columnwidth}{CCC}
\toprule
\textbf{Push (m/s)} & \textbf{DynaMITE} & \textbf{LSTM} \\
\midrule
1.0 & $3.96 \pm 0.53$ & $\mathbf{3.05 \pm 0.14}$ \\
2.0 & $4.03 \pm 0.37$ & $\mathbf{3.32 \pm 0.08}$ \\
3.0 & $4.36 \pm 0.39$ & $\mathbf{3.78 \pm 0.08}$ \\
4.0 & $5.01 \pm 0.26$ & $\mathbf{4.62 \pm 0.07}$ \\
5.0 & $5.76 \pm 0.22$ & $\mathbf{5.50 \pm 0.11}$ \\
6.0 & $6.74 \pm 0.17$ & $\mathbf{6.41 \pm 0.08}$ \\
8.0 & $8.45 \pm 0.15$ & $\mathbf{8.22 \pm 0.12}$ \\
\bottomrule
\end{tabularx}
\end{table}

However, LSTM achieves lower peak tracking error at all magnitudes (Table~\ref{tab:recovery_peak}). Post-push reward is also substantially better for LSTM ($-3.2$ to $-4.8$ vs.\ DynaMITE's $-9.5$ to $-9.8$). The result is narrower than ``better recovery'': under our protocol and adopted threshold, DynaMITE returns below the tracking criterion fastest but with worse peak displacement and worse overall locomotion quality; LSTM achieves the best gait but crosses the threshold slowest.

\subsection{Single-Axis OOD Sweeps}\label{sec:results_sweeps}

Table~\ref{tab:push_sweep} presents the push magnitude sweep on the randomized task. DynaMITE matches or exceeds LSTM from push level~4 (3--5\,m/s). Among these two models, LSTM degrades more steeply with increasing push magnitude (sensitivity 1.52 vs.\ 0.41).

\begin{table}[H]
\caption{Push magnitude sweep (randomized task, 5 seeds, 50-ep eval per level). Bold indicates best.\label{tab:push_sweep}}
\resizebox{\columnwidth}{!}{%
\begin{tabular}{lccccccc}
\toprule
\textbf{Method} & \textbf{0} & \textbf{0.5--1} & \textbf{1--2} & \textbf{2--3} & \textbf{3--5} & \textbf{5--8} & \textbf{S.} \\
\midrule
DynaMITE & $-4.37$ & $-4.44$ & $-4.50$ & $-4.56$ & $\mathbf{-4.63}$ & $\mathbf{-4.78}$ & 0.41 \\
LSTM & $\mathbf{-3.58}$ & $\mathbf{-4.05}$ & $\mathbf{-4.23}$ & $\mathbf{-4.45}$ & $-4.70$ & $-5.09$ & 1.52 \\
Transf. & $-4.65$ & $-4.72$ & $-4.76$ & $-4.81$ & $-4.94$ & $-5.08$ & 0.42 \\
MLP & $-5.65$ & $-5.64$ & $-5.79$ & $-5.75$ & $-5.89$ & $-5.97$ & $\mathbf{0.33}$ \\
\bottomrule
\end{tabular}%
}
\end{table}

Table~\ref{tab:degradation_summary} summarizes severe OOD degradation across all sweeps and tasks.

\begin{table}[H]
\caption{Severe OOD degradation across all sweep types and tasks (5 seeds). DynaMITE degrades less than LSTM on all push-related sweeps; crossover $=$ level where DynaMITE first matches or exceeds LSTM.\label{tab:degradation_summary}}
\small
\begin{tabularx}{\columnwidth}{CCCCC}
\toprule
\textbf{Sweep} & \textbf{Task} & \textbf{DynaMITE} & \textbf{LSTM} & \textbf{Cross.} \\
\midrule
Combined & rand. & $\mathbf{2.3\%}$ & 16.7\% & L3 \\
Push mag. & rand. & $\mathbf{5.1\%}$ & 17.1\% & L4 \\
Push mag. & push & $\mathbf{4.4\%}$ & 12.2\% & L5 \\
Push mag. & terrain & $\mathbf{6.6\%}$ & 21.5\% & L5 \\
Friction & rand. & $-0.4\%$\textsuperscript{*} & 1.5\% & None \\
\bottomrule
\end{tabularx}
\noindent{\footnotesize{\textsuperscript{*} Negative degradation indicates slight improvement under OOD conditions (within noise).}}
\end{table}

\begin{figure}[t]
\centering
\includegraphics[width=\columnwidth]{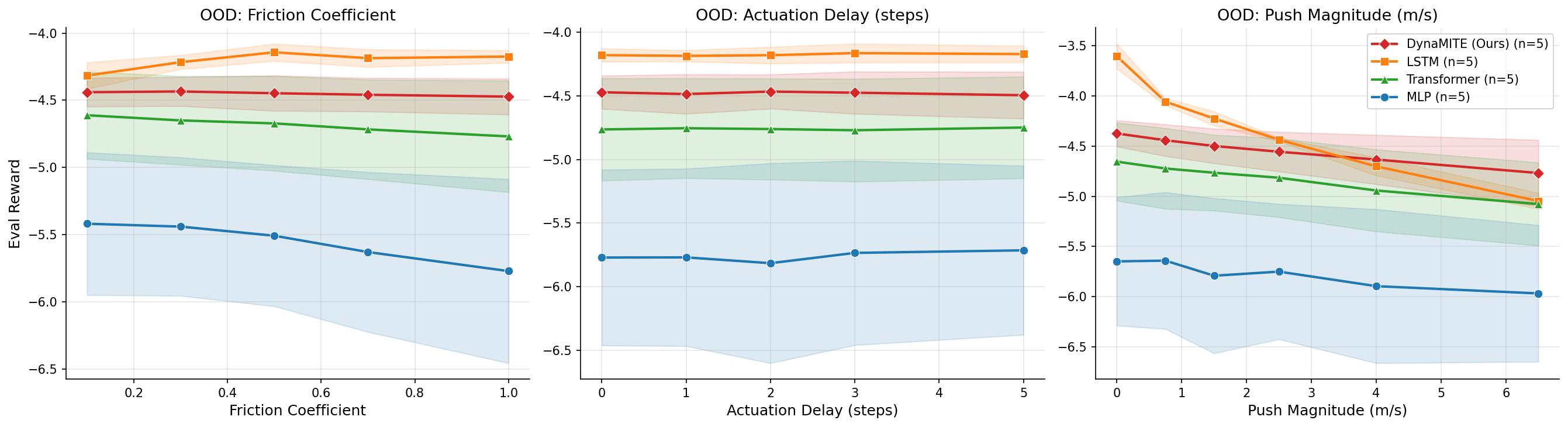}
\caption{OOD sweep comparison across four models and five seeds. LSTM degrades more steeply than DynaMITE under push magnitude perturbation across all three tasks.\label{fig:sweep_robustness}}
\end{figure}

Figure~\ref{fig:sweep_robustness} presents the full OOD sweep comparison. LSTM degrades more steeply than DynaMITE across all three tasks for push magnitude sweeps: sensitivity 1.52 (randomized), 1.41 (push), 1.61 (terrain) vs.\ DynaMITE's 0.39--0.46.

Only 3 of 42 pairwise OOD comparisons survive Holm--Bonferroni correction ($p_\text{adj} < 0.05$). Most ranking claims are directional, not statistically confirmed at $n = 5$.

\subsection{Probing and Intervention Do Not Support Decodable Factor Structure}\label{sec:results_repr}

\subsubsection{Factor Alignment}\label{sec:results_alignment}

\begin{figure}[t]
\centering
\includegraphics[width=0.7\columnwidth]{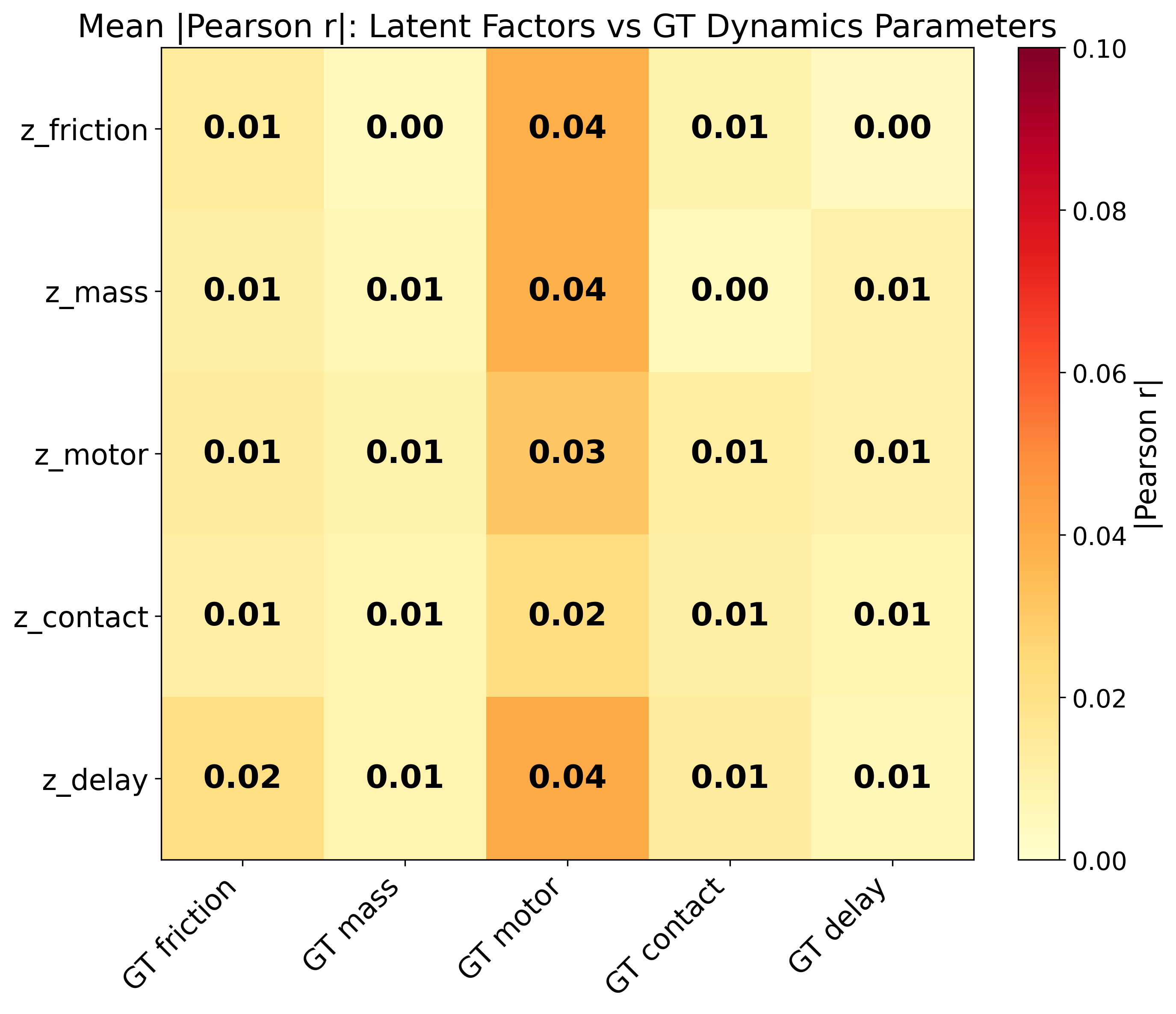}
\caption{Latent correlation heatmap showing the Pearson correlation between each factor subspace dimension and the corresponding ground-truth dynamics parameter.\label{fig:latent_heatmap}}
\end{figure}

On a custom within-factor correlation metric (chance level = 0.20 for five factors), DynaMITE achieves a mean alignment score of $0.500 \pm 0.020$ across three seeds. Figure~\ref{fig:latent_heatmap} shows the factor--subspace structure.

\subsubsection{Intervention Analysis (Negative Result)}\label{sec:results_intervention}

Table~\ref{tab:intervention} presents the reward change when clamping each factor subspace. All $|\Delta\text{reward}| < 0.05$ across 3 seeds $\times$ 5 factors $\times$ 3 DR levels (90 evaluations total). No evidence of factor-specific behavioral control.

\begin{table}[H]
\caption{Latent intervention analysis: average absolute reward change per factor subspace (3 seeds $\times$ 5 factors $\times$ 3 DR levels $=$ 90 conditions, 50-ep eval).\label{tab:intervention}}
\begin{tabularx}{\columnwidth}{CCC}
\toprule
\textbf{Factor (dims)} & \textbf{Avg $|\Delta\text{Reward}|$} & \textbf{Interpretation} \\
\midrule
Friction (0--3) & 0.007 & Negligible \\
Mass (4--9) & 0.012 & Negligible \\
Motor (10--15) & 0.021 & Negligible \\
Contact (16--19) & 0.020 & Negligible \\
Delay (20--23) & 0.020 & Negligible \\
\bottomrule
\end{tabularx}
\end{table}

\subsubsection{Latent Probe Analysis}\label{sec:results_probe}

We train Ridge regression (linear) and MLP (nonlinear) probes to predict ground-truth dynamics parameters from each model's internal representation (DynaMITE's 24-d factored latent vs.\ LSTM's 128-d hidden state), using ${\sim}36{,}000$ samples per run with 5-fold cross-validation (Table~\ref{tab:probe}).

\begin{table}[H]
\caption{Latent probe $R^2$ for DynaMITE vs.\ LSTM (5 seeds, 5-fold CV, $\sim$36k samples). Bold indicates best per factor.\label{tab:probe}}
\small
\begin{tabularx}{\columnwidth}{CCCCC}
\toprule
\textbf{Factor} & \textbf{DM Lin.} & \textbf{DM MLP} & \textbf{LSTM Lin.} & \textbf{LSTM MLP} \\
\midrule
Friction & $-0.000$ & $-0.001$ & $0.026$ & $\mathbf{0.101}$ \\
Mass & $0.000$ & $-0.002$ & $0.012$ & $\mathbf{0.018}$ \\
Motor & $0.000$ & $-0.002$ & $0.010$ & $\mathbf{0.015}$ \\
Contact & $0.000$ & $-0.000$ & $0.006$ & $\mathbf{0.045}$ \\
Delay & $0.000$ & $-0.000$ & $0.011$ & $\mathbf{0.041}$ \\
\midrule
\textbf{Overall} & \textbf{0.000} & \textbf{$-$0.001} & \textbf{0.013} & \textbf{0.044} \\
\bottomrule
\end{tabularx}
\end{table}

Despite being trained with per-factor auxiliary losses, DynaMITE's latent shows $R^2 \approx 0$ for both linear and nonlinear probes across all five factors. LSTM's hidden state, with no auxiliary signal, achieves $R^2$ up to 0.101 (friction, MLP probe). However, absolute $R^2$ values are low for both models (LSTM overall: 0.044).

\subsection{Ablation Study}\label{sec:results_ablation}

To assess the contribution of each architectural component, we expand the ablation study to $n = 10$ seeds (42--51) per variant on the randomized task. Table~\ref{tab:ablation} reports the results. The \textbf{Single Latent} variant---which replaces the five-factor partition with a single 24-d unfactored latent trained with a monolithic auxiliary loss---reaches statistical significance ($p = 0.048$), with 8 of 10 seeds degrading (mean $\Delta = -0.23$). No Latent shows consistent degradation (9/10 seeds worse, $\Delta = -0.18$) but does not reach significance ($p = 0.225$) due to high variance. No Aux Loss remains directionally negative but inconsistent (6/10 seeds worse, $p = 0.641$). The \textbf{Aux Only} variant---auxiliary losses active but the policy sees only the full 128-d transformer features---performs identically to No Latent ($-4.64$ vs.\ $-4.64$, $p = 0.251$).

\begin{table}[H]
\caption{Ablation study on the randomized task (10 seeds, paired $t$-test vs.\ full DynaMITE). Bold $p$-value indicates $p < 0.05$.\label{tab:ablation}}
\small
\begin{tabularx}{\columnwidth}{CCCCC}
\toprule
\textbf{Variant} & \textbf{Reward} & \textbf{$\Delta$} & \textbf{$p$} & \textbf{Worse} \\
\midrule
Full & $\mathbf{-4.46}$ & --- & --- & --- \\
No Aux & $-4.51$ & $-0.05$ & 0.641 & 6/10 \\
No Latent & $-4.64$ & $-0.18$ & 0.225 & 9/10 \\
Single & $-4.69$ & $-0.23$ & $\mathbf{0.048}$ & 8/10 \\
Aux Only & $-4.64$ & $-0.18$ & 0.251 & 7/10 \\
\bottomrule
\end{tabularx}
\end{table}

\subsubsection{Factorial: Bottleneck vs.\ Aux Loss}\label{sec:results_factorial}

To decompose whether the observed reward difference comes from the $\tanh$ bottleneck, the auxiliary losses, or both, we add a fourth ablation cell---\textbf{Aux Only}: the latent head and per-factor auxiliary losses are active (providing gradient regularization to the transformer), but the policy and value heads see only the full 128-d transformer features, not the 24-d compressed bottleneck. Table~\ref{tab:factorial} presents the $2 \times 2$ factorial.

\begin{table}[H]
\caption{$2 \times 2$ factorial: bottleneck vs.\ auxiliary loss (10 seeds, randomized task). Each cell averages 10 paired seeds.\label{tab:factorial}}
\begin{tabularx}{\columnwidth}{CCC}
\toprule
 & \textbf{No Aux Loss} & \textbf{Aux Loss} \\
\midrule
\textbf{Bottleneck} & $-4.51 \pm 0.20$ & $\mathbf{-4.46 \pm 0.25}$ \\
\textbf{No Bottleneck} & $-4.64 \pm 0.36$ & $-4.64 \pm 0.34$ \\
\bottomrule
\end{tabularx}
\end{table}

Factorial main effects (paired $t$-tests, $n = 10$): the auxiliary losses show \textbf{no measurable effect} on ID reward ($+0.03$, $p = 0.732$); the bottleneck shows a \textbf{consistent advantage} ($+0.16$, $p = 0.207$); the interaction is negligible ($+0.05$, $p = 0.787$).

\subsubsection{Factorial: OOD Robustness}\label{sec:results_factorial_ood}

To test whether the bottleneck's advantage amplifies under distribution shift, we evaluate all four $2 \times 2$ cells on \textbf{combined-shift severe} (Levels~3--4). Table~\ref{tab:factorial_ood} presents the severe OOD results.

\begin{table}[H]
\caption{$2 \times 2$ factorial under severe combined-shift (avg.\ of Levels~3--4, 10 seeds, 50-ep eval per level).\label{tab:factorial_ood}}
\begin{tabularx}{\columnwidth}{CCC}
\toprule
 & \textbf{No Aux Loss} & \textbf{Aux Loss} \\
\midrule
\textbf{Bottleneck} & $-4.59 \pm 0.21$ & $\mathbf{-4.55 \pm 0.22}$ \\
\textbf{No Bottleneck} & $-4.68 \pm 0.28$ & $-4.66 \pm 0.23$ \\
\bottomrule
\end{tabularx}
\end{table}

OOD factorial main effects: auxiliary losses show \textbf{no measurable effect} ($+0.034$, $p = 0.669$); the bottleneck advantage \textbf{persists but does not amplify} ($+0.100$, $p = 0.208$); interaction is absent ($+0.015$, $p = 0.912$). Table~\ref{tab:factorial_degradation} shows that all four variants are remarkably robust under severe perturbation (degradation $0.5$--$2.1\%$), and that bottleneck models degrade slightly \emph{more} than no-bottleneck variants.

\begin{table}[H]
\caption{Factorial degradation analysis (10 seeds, randomized task). Degradation $= |\text{Severe} - \text{ID}|/|\text{ID}|$.\label{tab:factorial_degradation}}
\small
\begin{tabularx}{\columnwidth}{CCCCC}
\toprule
\textbf{Variant} & \textbf{ID} & \textbf{Severe} & \textbf{$\Delta$} & \textbf{Degrad.} \\
\midrule
Full (B+A) & $-4.46$ & $-4.55$ & $-0.09$ & $2.1\%$ \\
No Aux (B) & $-4.51$ & $-4.59$ & $-0.08$ & $1.9\%$ \\
No Latent & $-4.64$ & $-4.68$ & $-0.04$ & $0.9\%$ \\
Aux Only (A) & $-4.64$ & $-4.66$ & $-0.02$ & $0.5\%$ \\
\bottomrule
\end{tabularx}
\end{table}

\subsection{Summary of Evidence}\label{sec:results_summary}

Table~\ref{tab:evidence_summary} summarizes the key claims and their evidence status.

\begin{table}[H]
\caption{Summary of key claims and evidence status.\label{tab:evidence_summary}}
\footnotesize
\begin{tabularx}{\columnwidth}{CCC}
\toprule
\textbf{Claim} & \textbf{Evidence} & \textbf{Status} \\
\midrule
Latent decodable? & Probe $R^2 \approx 0$; MIG/DCI/SAP $\approx 0$ & \textbf{No} \\
Factor-specific effect? & $|\Delta r| < 0.05$ & No evidence \\
Aux heads learn targets? & MSE $\approx 3.7$ (48\% drop) & Not accurately \\
LSTM best ID? & $p < 0.03$, all 4 tasks & \textbf{Yes} \\
DynaMITE lower OOD sens.? & 0.25 vs.\ 1.57 & Directional \\
DynaMITE faster recovery? & ${\sim}6$ vs.\ 9--21 steps & Directional \\
Regularization mechanism? & Gradient orth., low rank, MI & Consistent \\
Bottleneck drives reward? & $+0.16$ ($p = 0.207$) & Consistent \\
Bottleneck drives OOD? & $+0.10$ ($p = 0.208$) & Consistent \\
\bottomrule
\end{tabularx}
\end{table}

\section{Mechanistic Analysis}\label{sec:mechanistic}

To investigate potential mechanisms behind DynaMITE's less pronounced OOD degradation despite the absence of a decodable dynamics representation, we examine four aspects of the training and representation: gradient flow, representation geometry, mutual information, and standard disentanglement metrics.

\subsection{Gradient Flow Analysis}\label{sec:mech_gradient}

We retrain DynaMITE (3 seeds, 10M steps each) with gradient instrumentation, computing separate gradient vectors for the PPO objective and each of the five auxiliary factor losses at every 10th iteration (${\sim}81$ data points per seed). Table~\ref{tab:cosine} reports the cosine similarity between PPO and auxiliary gradients.

\begin{table}[H]
\caption{Cosine similarity between PPO and auxiliary gradients (3 seeds, averaged over training).\label{tab:cosine}}
\begin{tabularx}{\columnwidth}{CCC}
\toprule
\textbf{Factor} & \textbf{Mean $\cos(\nabla_\text{PPO}, \nabla_\text{aux})$} & \textbf{Std} \\
\midrule
Friction & $-0.001$ & 0.005 \\
Mass & $-0.002$ & 0.010 \\
Motor & $-0.005$ & 0.016 \\
Contact & $-0.008$ & 0.026 \\
Delay & $-0.002$ & 0.009 \\
\bottomrule
\end{tabularx}
\end{table}

\begin{figure}[t]
\centering
\includegraphics[width=\columnwidth]{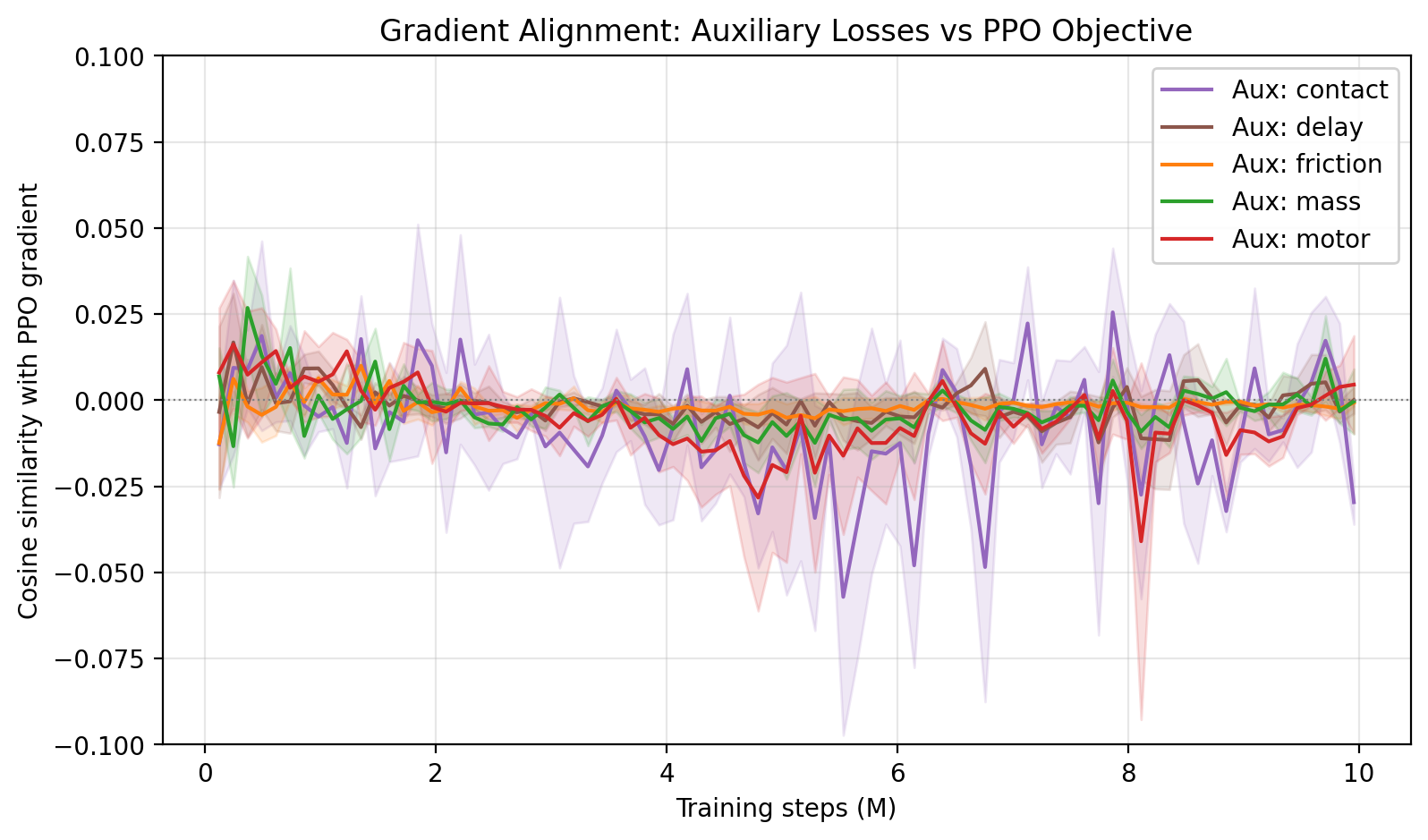}
\caption{Cosine similarity between PPO and auxiliary gradients over training (3 seeds). All five factor-specific auxiliary gradients remain orthogonal to the PPO gradient ($|\cos| < 0.01$) throughout the entire training run.\label{fig:cosine_sim}}
\end{figure}

All cosine similarities are indistinguishable from zero ($|\cos| < 0.01$), as shown in Figure~\ref{fig:cosine_sim}. The auxiliary gradients are orthogonal to the PPO gradient throughout training. They contribute approximately 20--40\% of the total gradient norm but in orthogonal directions, consistent with a regularization-like effect rather than direct optimization synergy.

\textbf{Auxiliary head convergence.} The total auxiliary MSE drops from $7.2 \pm 0.7$ at the start of training to $3.7 \pm 0.1$ at convergence (5 seeds), a 48\% reduction. However, the residual MSE remains high (per-factor average $\approx 0.75$), indicating that the auxiliary heads do not learn to accurately predict their target dynamics parameters.

\subsection{Representation Geometry}\label{sec:mech_geometry}

We collect ${\sim}36{,}000$ representations per model/seed (200 episodes $\times$ 32 environments) from trained checkpoints and compute SVD-based geometry metrics (Table~\ref{tab:geometry}).

\begin{table}[H]
\caption{Representation geometry comparison (5 seeds, SVD on $\sim$36k representations per model).\label{tab:geometry}}
\begin{tabularx}{\columnwidth}{CCC}
\toprule
\textbf{Metric} & \textbf{DynaMITE (24-d)} & \textbf{LSTM (128-d)} \\
\midrule
Effective rank & $\mathbf{4.78 \pm 0.72}$ & $32.20 \pm 3.85$ \\
Participation ratio & $\mathbf{2.27 \pm 0.15}$ & $4.96 \pm 1.52$ \\
Condition number & $315.9 \pm 204.1$ & $\mathbf{108.4 \pm 19.9}$ \\
\bottomrule
\end{tabularx}
\end{table}

DynaMITE's latent is much lower-dimensional in practice: effective rank ${\sim}5$ out of 24 possible dimensions, compared with LSTM's ${\sim}32$ out of 128. The high condition number (316 vs.\ 108) indicates a highly anisotropic structure.

\begin{figure}[t]
\centering
\includegraphics[width=0.85\columnwidth]{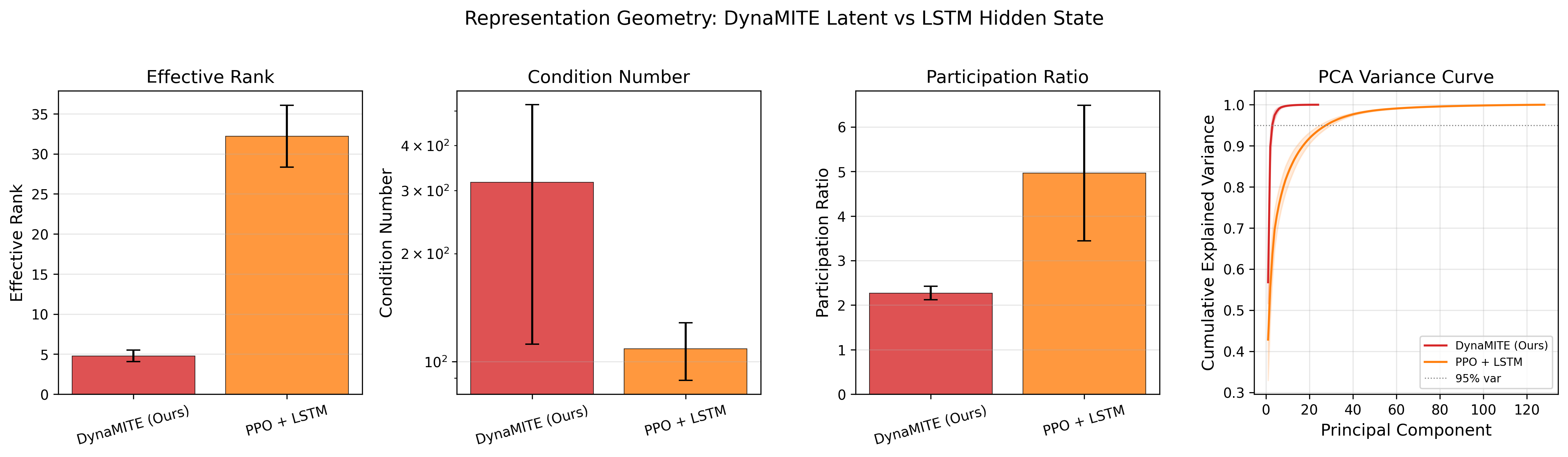}
\caption{Representation geometry. DynaMITE's 24-d latent collapses to effective rank ${\sim}5$; LSTM's 128-d hidden state uses ${\sim}32$ effective dimensions.\label{fig:geometry}}
\end{figure}

\subsection{Mutual Information Estimation}\label{sec:mech_mi}

We estimate mutual information (MI) between learned representations and ground-truth dynamics parameters using KNN estimation (\texttt{sklearn mutual\_info\_regression}, $k = 5$). Table~\ref{tab:mi} reports the results.

\begin{table}[H]
\caption{Mutual information between representations and dynamics factors (5 seeds, KNN estimation $k=5$, $\sim$36k samples, nats).\label{tab:mi}}
\begin{tabularx}{\columnwidth}{CCC}
\toprule
\textbf{Factor} & \textbf{DynaMITE (nats)} & \textbf{LSTM (nats)} \\
\midrule
\textbf{Overall} & $\mathbf{0.233 \pm 0.052}$ & $0.028 \pm 0.033$ \\
Friction & $\mathbf{0.054 \pm 0.021}$ & $0.024 \pm 0.020$ \\
Mass & $\mathbf{0.091 \pm 0.010}$ & $0.011 \pm 0.013$ \\
Motor & $\mathbf{0.037 \pm 0.031}$ & $0.013 \pm 0.015$ \\
Contact & $\mathbf{0.026 \pm 0.011}$ & $0.010 \pm 0.015$ \\
Delay & $0.007 \pm 0.004$ & $0.005 \pm 0.003$ \\
\bottomrule
\end{tabularx}
\end{table}

DynaMITE's latent contains more MI with dynamics parameters than LSTM's hidden state (0.233 vs.\ 0.028 nats overall). However, all MI values are small in absolute terms ($< 0.25$ nats).

\begin{figure}[t]
\centering
\includegraphics[width=0.85\columnwidth]{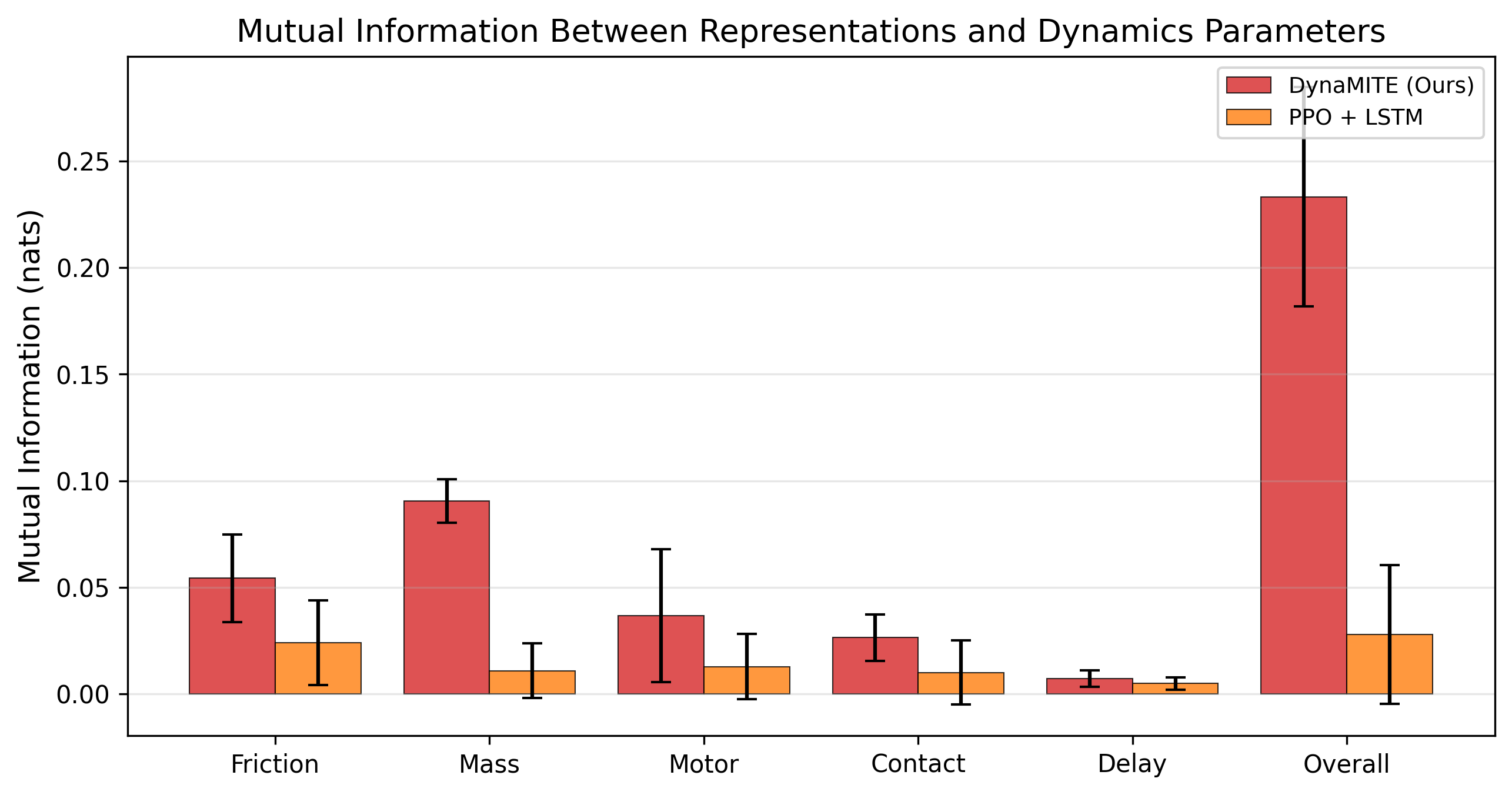}
\caption{Mutual information between learned representations and ground-truth dynamics factors (KNN estimation). DynaMITE retains more MI with dynamics parameters than LSTM (0.233 vs.\ 0.028 nats), but absolute values for both models are low.\label{fig:mi_comparison}}
\end{figure}

\subsection{Standard Disentanglement Metrics}\label{sec:mech_disentangle}

We compute MIG~\cite{mig}, DCI~\cite{dci}, and SAP~\cite{sap} on the same latent representations (Table~\ref{tab:disentangle}).

\begin{table}[H]
\caption{Standard disentanglement metrics (5 seeds). Lower DCI-I is better (prediction error).\label{tab:disentangle}}
\begin{tabularx}{\columnwidth}{CCC}
\toprule
\textbf{Metric} & \textbf{DynaMITE} & \textbf{LSTM} \\
\midrule
MIG & $0.0013$ & $0.0013$ \\
DCI Disent. & $0.0195$ & $\mathbf{0.0929}$ \\
DCI Complete & $0.0168$ & $\mathbf{0.0542}$ \\
DCI Info. & $\mathbf{0.0846}$ & $0.1397$ \\
SAP & $0.0001$ & $0.0006$ \\
\bottomrule
\end{tabularx}
\end{table}

All scores are very low for both models---neither architecture produces cleanly disentangled representations. The near-zero MIG and SAP for both models confirm that no individual latent dimension aligns with a single dynamics factor.

\subsection{Mechanistic Synthesis}\label{sec:mech_synthesis}

We separate the findings into four categories:

\textbf{Observed facts.} (i)~Auxiliary and PPO gradients are orthogonal throughout training ($|\cos| < 0.01$). (ii)~The 24-d latent collapses to effective rank ${\sim}5$. (iii)~The compressed latent retains more MI with dynamics than LSTM's hidden state (0.233 vs.\ 0.028 nats), though absolute MI is low for both. (iv)~Neither representation is probe-decodable. (v)~Standard disentanglement metrics (MIG, DCI, SAP) are near zero for both models; LSTM is ${\sim}5\times$ more disentangled per DCI.

\textbf{Plausible interpretation.} The auxiliary losses are consistent with a regularization-like effect that constrains features into a low-dimensional subspace. This compression could contribute to less OOD degradation by limiting the representational capacity available for overfitting to the training distribution, but this interpretation is speculative.

\textbf{Not established.} Whether compression causally mediates the observed OOD sensitivity difference. The $2 \times 2$ factorial identifies the $\tanh$ bottleneck---not the auxiliary losses---as the component driving the reward advantage, but neither main effect reaches significance ($p \approx 0.2$). Architectural capacity differences and optimization landscape effects remain viable alternative explanations.

\section{Discussion}\label{sec:discussion}

\subsection{Observed Operating Regimes}\label{sec:disc_when}

The results suggest an ID--OOD tradeoff under our evaluation protocol (Table~\ref{tab:guidelines}).

\begin{table}[H]
\caption{Observed patterns under our simulation protocol.\label{tab:guidelines}}
\begin{tabularx}{\columnwidth}{CCC}
\toprule
\textbf{Scenario} & \textbf{Favored} & \textbf{Evidence} \\
\midrule
Nominal ID reward & LSTM & Confirmed \\
Moderate mismatch & LSTM & Directional \\
Severe mismatch & DynaMITE & Directional \\
Fast re-adaptation & DynaMITE & Directional \\
Inspect latent & Neither & Confirmed \\
\bottomrule
\end{tabularx}
\end{table}

\subsection{Practical Takeaway}\label{sec:disc_practical}

\begin{itemize}[leftmargin=*]
    \item \textbf{For nominal reward:} LSTM remains the strongest choice. It achieves the best ID reward on all tasks and degrades gracefully under moderate perturbation.
    \item \textbf{For severe OOD robustness:} DynaMITE shows less degradation than LSTM under combined shift (2.3\% vs.\ 16.7\%), but the $2 \times 2$ factorial attributes this to the bottleneck compression, not the auxiliary supervision.
    \item \textbf{For interpretability:} Auxiliary dynamics supervision does not produce latent representations useful for online dynamics monitoring or debugging.
    \item \textbf{Mechanism:} The bottleneck is the primary component driving observed reward differences. The auxiliary losses show no measurable contribution when the bottleneck is held constant ($p > 0.66$).
\end{itemize}

\subsection{Why Does LSTM Degrade More Under Combined Shift?}\label{sec:disc_why}

LSTM degrades more than DynaMITE under combined perturbation. The $2 \times 2$ factorial clarifies this: within the DynaMITE architecture, the bottleneck shows a consistent advantage under severe combined shift ($+0.10$, $p = 0.208$), while the auxiliary losses show none ($+0.03$, $p = 0.669$). Crucially, bottleneck models degrade slightly \emph{more} than no-bottleneck variants ($2.1\%$ vs.\ $0.5\%$, Table~\ref{tab:factorial_degradation}), confirming the advantage is a training-time benefit that carries through to OOD rather than a specific robustness mechanism. The larger DynaMITE-vs.-LSTM gap (2.3\% vs.\ 16.7\%) therefore reflects architectural differences between the two model families, not the auxiliary supervision.

\section{Limitations}\label{sec:limitations}

\begin{itemize}[leftmargin=*]
    \item \textbf{Partially deconfounded comparison.} The $2 \times 2$ factorial identifies the bottleneck as the primary component, but neither main effect reaches significance ($p \approx 0.2$).
    \item \textbf{Limited statistical power.} Only 3 of 42 pairwise OOD comparisons survive Holm--Bonferroni correction at $n = 5$. Most OOD claims are directional.
    \item \textbf{Probe family.} Our probes are Ridge regression and small MLPs (1 hidden layer, 64 units). More expressive decoders might recover information our probes miss.
    \item \textbf{Sensitivity metric.} The OOD sensitivity metric ($\max - \min$) does not distinguish genuine robustness from uniformly poor performance.
    \item \textbf{Simulation only.} All experiments use Isaac Lab. Hardware transfer is untested.
    \item \textbf{Deterministic evaluation only.} Stochastic policy behavior is unreported.
    \item \textbf{Narrow reward spread.} The top two models span ${\sim}0.6$ reward units; practical significance for deployment is unknown.
    \item \textbf{Incomplete perturbation coverage.} Mass, contact stiffness, and observation noise are untested as single-axis sweeps.
\end{itemize}

\paragraph{What this paper does not claim.}
This paper does not show that decodable or disentangled latent dynamics are impossible in locomotion---only that one specific supervision strategy did not produce them in this setup. It does not establish real-world transfer. It does not claim that auxiliary losses are without value; they may act as useful regularizers even when they fail to produce interpretable latent structure.

\section{Future Work}\label{sec:future}

\begin{itemize}[leftmargin=*]
    \item \textbf{Sim-to-real validation.} Hardware deployment on a Unitree G1 to test whether the OOD robustness tradeoff transfers.
    \item \textbf{Extending the factorial to push recovery.} The $2 \times 2$ factorial has not yet been applied to push-recovery protocols. Additionally, varying bottleneck width could further clarify the role of compression.
    \item \textbf{Advanced disentanglement methods.} Nonlinear ICA or manifold-aware probes may recover structure invisible to standard metrics.
    \item \textbf{Increased statistical power.} Expanding seeds beyond $n = 10$ (e.g., $n = 20$) with bootstrap confidence intervals.
\end{itemize}

\section{Conclusions}\label{sec:conclusions}

We evaluated whether factor-wise auxiliary dynamics supervision produces decodable or functionally separable latent structure in simulated humanoid locomotion. Under the probes, interventions, and disentanglement metrics adopted here, it does not: probes yield $R^2 \approx 0$, clamping produces negligible reward change, and standard metrics are near zero. An unsupervised LSTM hidden state achieves higher probe accuracy.

A $2 \times 2$ factorial ablation ($n = 10$) isolates the contributions of the $\tanh$ bottleneck and auxiliary losses. The observed performance differences are attributable primarily to the bottleneck compression, not the auxiliary supervision: the auxiliary losses show no measurable effect on either ID reward ($+0.03$, $p = 0.732$) or severe OOD reward ($+0.03$, $p = 0.669$), while the bottleneck yields a consistent advantage in both regimes (ID: $+0.16$, $p = 0.207$; OOD: $+0.10$, $p = 0.208$). The bottleneck's advantage persists under severe perturbation but does not amplify, and bottleneck models degrade slightly more than no-bottleneck variants---confirming a training-time representation benefit rather than a robustness-specific mechanism.

For practitioners: LSTM remains the strongest choice when nominal performance is the priority. Auxiliary dynamics supervision does not produce an interpretable estimator and does not measurably improve reward beyond what the bottleneck alone provides. The value of multi-component architectures like DynaMITE, if any, lies in the implicit compression imposed by the bottleneck, not in the explicit supervision signal.

\section*{Data Availability}

All training scripts, evaluation code, configuration files, and raw results are publicly available at \url{https://github.com/fjkrch/g1-factorized-latent-locomotion}. No pre-trained checkpoints are provided; all models must be trained from scratch using the released code and configurations. Full reproduction from source requires approximately 35 hours on a single NVIDIA RTX 4060 GPU.

\section*{Acknowledgments}

During the preparation of this manuscript, the author used GitHub Copilot (powered by Claude, Anthropic) for the purposes of drafting and editing selected sections. The author has reviewed and edited the output and takes full responsibility for the content of this publication.

\bibliographystyle{unsrt}

\end{document}